\documentclass[10pt,twocolumn,letterpaper]{article}

\usepackage[accsupp]{axessibility} 
\usepackage{iccv}
\usepackage{times}
\usepackage{epsfig}
\usepackage{graphicx}
\usepackage{url}

\usepackage{amsmath}
\usepackage{amssymb}
\usepackage{microtype}
\usepackage{graphicx}
\usepackage{booktabs} % for professional tables
\usepackage{lipsum}
\usepackage{amsmath}
\usepackage{adjustbox}
\usepackage{algorithm}
\usepackage{algorithmic}
\usepackage{graphicx}
\usepackage{subcaption}
\usepackage{authblk}
\graphicspath{ {./images/} }

\input{Definitions}

\usepackage[pagebackref=true,breaklinks=true,letterpaper=true,colorlinks,bookmarks=false]{hyperref}

\iccvfinalcopy % *** Uncomment this line for the final submission

\def\iccvPaperID{9659} % *** Enter the ICCV Paper ID here

\ificcvfinal\fi

\begin{document}

%%%%%%%%% TITLE
\title{Meta Learning on a Sequence of Imbalanced Domains with Difficulty Awareness}

\author[1]{Zhenyi Wang }
\author[1]{Tiehang Duan }
\author[1]{Le Fang}
\author[1]{Qiuling Suo}
\author[1]{Mingchen Gao}
\affil[1]{Department of Computer Science and Engineering, University at Buffalo, SUNY}
% \affil[2]{Department of Mechanical Engineering, \LaTeX\ University}

\affil[1]{\textit {\{zhenyiwa, tiehangd, lefang, qiulings, mgao8\}@buffalo.edu}}

\renewcommand\Authands{ and }

\maketitle
% Remove page # from the first page of camera-ready.
\ificcvfinal\thispagestyle{empty}\fi

%%%%%%%%% ABSTRACT
\begin{abstract}
  Recognizing new objects by learning from a few labeled examples in an evolving environment is crucial to obtain excellent generalization ability for real-world machine learning systems. A typical setting across current meta learning algorithms assumes a stationary task distribution during meta training. In this paper, we explore a more practical and challenging setting where task distribution changes over time with domain shift. Particularly, we consider realistic scenarios where task distribution is highly imbalanced with domain labels unavailable in nature. We propose a kernel-based method for domain change detection and a difficulty-aware memory management mechanism that jointly considers the imbalanced domain size and domain importance to learn across domains continuously. Furthermore, we introduce an efficient adaptive task sampling method during meta training, which significantly reduces task gradient variance with theoretical guarantees. Finally, we propose a challenging benchmark with imbalanced domain sequences and varied domain difficulty. We have performed extensive evaluations on the proposed benchmark, demonstrating the effectiveness of our method. We made our code publicly available at \url{https://github.com/joey-wang123/Imbalancemeta.git}. 
\end{abstract}

%%%%%%%%% BODY TEXT
\section{Introduction}

Learning from a few labeled examples to acquire skills for new task is essential to achieve machine intelligence.  Take object recognition in personalized self-driving system as an example \cite{bae2020selfdriving}. Learning each user's personal driving preference model forms one task. The system is first deployed in the small city, Rochester. The company later extends its market to New York. The user base of New York is much larger than that of Rochester, causing domain imbalance. Also, after adapting to New York users, the learned user behavior from Rochester will be easily forgotten. Similar scenarios occur when learning to solve NLP tasks on a sequence of different languages \cite{incrementNLP} with imbalanced resources of different languages.

\begin{figure} 
     \centering
     \begin{subfigure}[b]{0.5\textwidth}
         \centering
         \includegraphics[width=\textwidth]{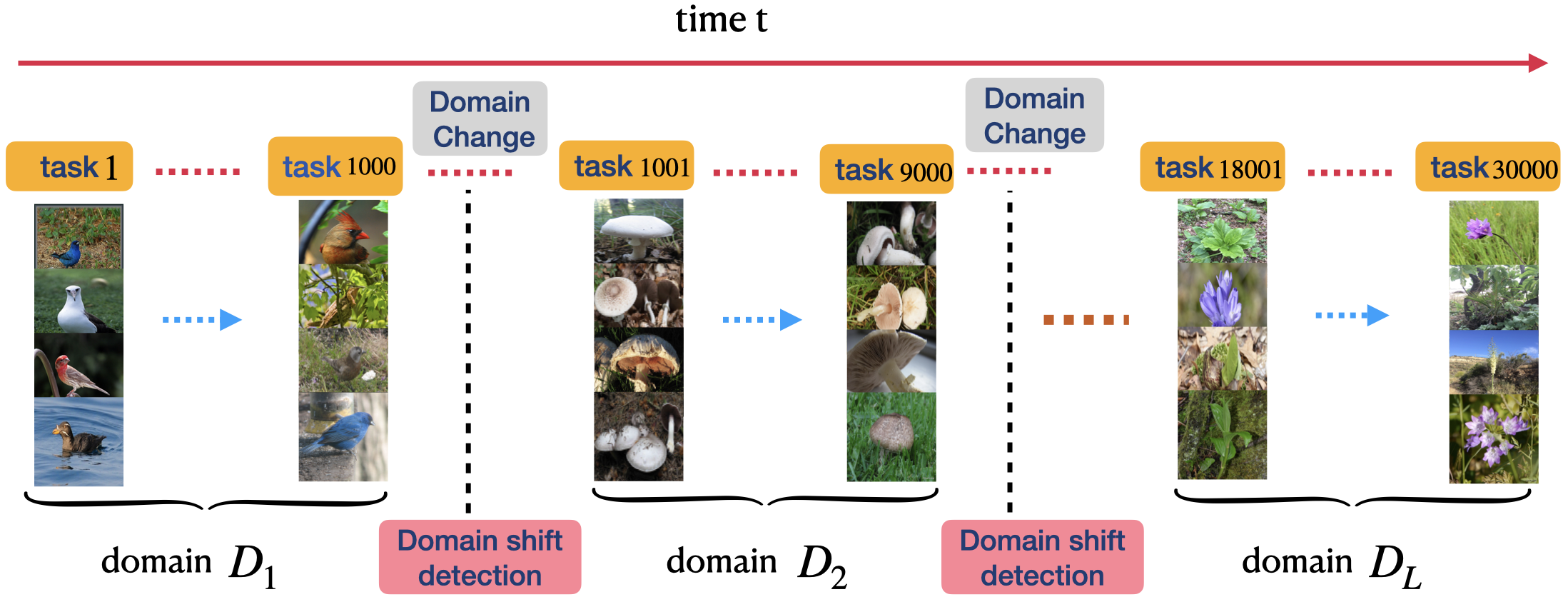}

     \end{subfigure}
     \hfill
     \begin{subfigure}[b]{0.5\textwidth}
         \centering
         \includegraphics[width=\textwidth]{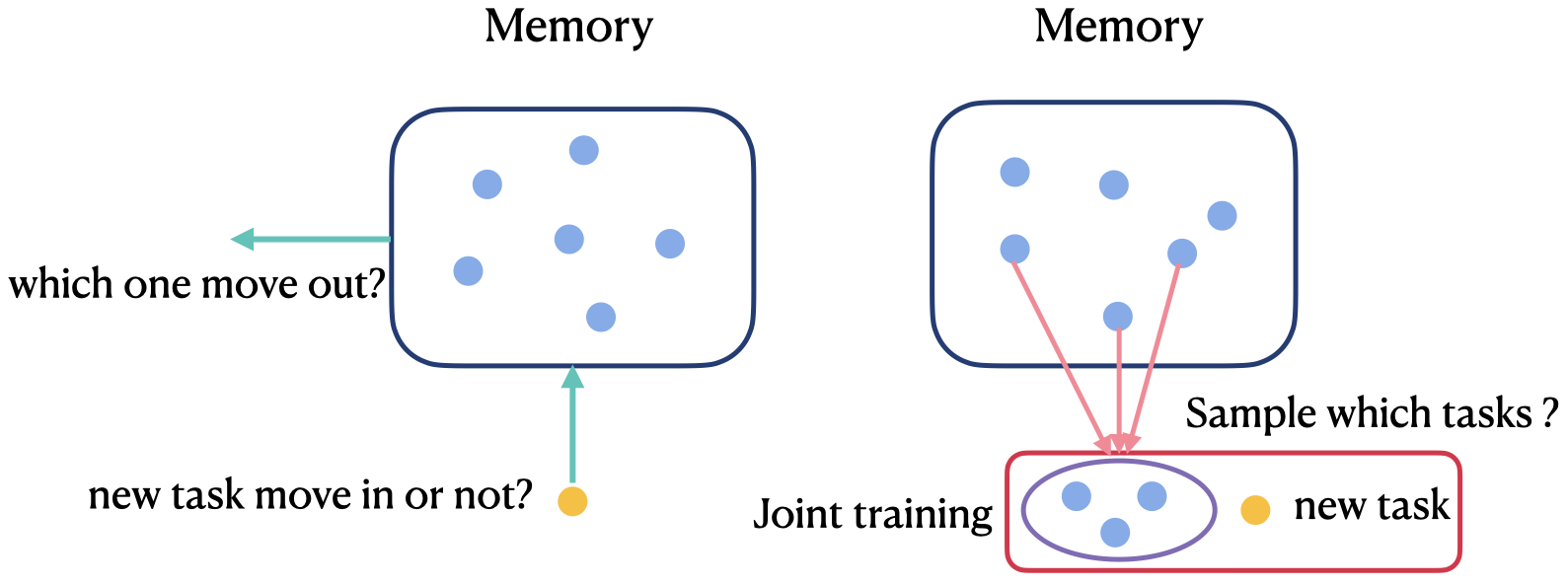}
        
     \end{subfigure}
     \caption{Illustration of meta learning for few shot object recognition on a sequence of imbalanced domains. Our focused problems including domain change detection, how to manage memory and sample memory tasks for joint training with streaming tasks.}
    \label{fig:overview}
\end{figure}

Meta learning is a promising approach for solving such few-shot learning problems. One common assumption of current models is that the task distribution is stationary during meta training. 
However, real world scenarios (such as the above self-driving system) are more complex and often involve learning across different domains (environments),  with challenges such as: (1) task distributions change among different domains; (2) tasks from previous domains are usually unavailable when training on a new domain; (3) the number of tasks from each domain could be highly imbalanced; (4) domain difficulty could vary significantly in nature across the domain sequence. An example is shown in Figure \ref{fig:overview}. Directly applying current meta learning models to such scenarios is not suitable to tackle these challenges, e.g., the object recognition accuracy of meta learned neural networks generally deteriorates significantly on previous context after adapting to a new environment \cite{gidaris2018dynamic, ren2018incremental, yoon2020xtarnet}.

In this work, we cope with such challenges by considering a more realistic problem setting that (1) learning on a sequence of domains; (2) task stream contains significant domain size imbalance; (3) domain labels and boundaries remain unavailable during both training and testing; (4) domain difficulty is non-uniform across the domain sequence. We term such problem setup as \textit{Meta Learning on a Sequence of Imbalanced Domains with Varying Difficulty} (MLSID).

MLSID requires the meta learning model both adapting to a new domain and retaining the ability to recognize objects from previous domains. To tackle this challenging problem, we adopt replay-based approaches, i.e., a small number of tasks from previous domains are maintained in a memory buffer. Accordingly, there are two main problems that need to be solved: (1) how to determine which task should be stored into the memory buffer and which to be moved out. To address this problem, we propose an adaptive memory management mechanism based on the domain distribution and difficulty, so that the tasks in memory buffer could maximize the retained knowledge of previous domains; (2) how to determine which tasks to sample from memory during meta training. We propose an efficient adaptive task sampling approach to accelerate meta training and reduce gradient estimation variance according to our derived optimal task sampling distribution. Our intuition is that not all tasks are equally important for joint training at different iterations. It is thus desirable to dynamically determine which tasks to sample and to be jointly trained with current tasks to mitigate catastrophic forgetting at each training iteration.

Our contributions are summarized as following:

\begin{itemize}
    \item To our best knowledge, this is the first work of meta learning on a sequence of imbalanced domains. For convenient evaluation of different models, we propose a new challenging benchmark consisting of imbalanced domain sequences.
    \item We propose a novel mechanism, \textit{``Memory Management with Domain Distribution and Difficulty Awareness''}, to maximize the retained knowledge of previous domains in the memory buffer. 
    \item We propose an efficient adaptive task sampling method during meta training, which significantly reduces gradient estimation variance with theoretical guarantees, making the meta training process more stable and boosting the model performance.
    \item Our method is orthogonal to specific meta learning methods and can be integrated with them seamlessly. Extensive experiments with gradient-based and metric-based meta learning methods on the proposed benchmark demonstrate the effectiveness of our method. 
\end{itemize}

\section{Problem Setting}

 A series of mini-batch training tasks ${\mathcal{T}}_{1}, {\mathcal{T}}_{2}, \dots, {\mathcal{T}}_{N}$ arrive sequentially, with possible domain shift occurring in the stream, i.e., the task stream can be  segmented by continual latent domains, ${\mathcal{D}}_{1}, {\mathcal{D}}_{2}, \dots, {\mathcal{D}}_{L}$.  $\mathcal{T}_{t}$ denotes the mini-batch of tasks arrived at time $t$. The domain identity associated with each task remains unavailable during both meta training and testing. Domain boundaries, i.e., indicating current domain has finished
and the next domain is about to start, are unknown. This is a more practical and general setup. Each task $\mathcal{T}$ is divided into training and testing data $\{\mathcal{T}^{train}, \mathcal{T}^{test} \}$. Suppose $\mathcal{T}_t^{train}$ consists of $K$ examples,  $\{(\bm{x}^k, \bm{y}^k)\}_{k=1}^{K}$, where in object recognition, $\bm{x}^k$ is the image data and $\bm{y}^k$ is the corresponding object label. 
We assume the agent stays within each domain for some consecutive time. Also, we consider a simplified setting where the agent will not return back to previous domains and put the contrary case into future work. Our proposed learning system maintains a memory buffer $\mathcal{M}$ to store a small number of training tasks from previous domains for replay to avoid forgetting of previous knowledge. Old tasks are not revisited during training unless they are stored in the memory $\mathcal{M}$. The total number of tasks processed is much larger than memory capacity.
At the end of meta training, we randomly sample a large number of \textit{unseen few-shot tasks} from each latent domain, ${\mathcal{D}}_{1}, {\mathcal{D}}_{2}, \dots, {\mathcal{D}}_{L}$ for meta testing. The model performance is the average accuracy on all the sampled tasks.

\section{Methodology}

\subsection{Conventional Reservoir Sampling and Its Limitations}
Reservoir sampling (RS) \cite{reservoir1985, ERRing19} is a random sampling method for choosing $k$ samples from a data stream in a single pass without knowing the actual value of total number of items in advance.
Straightforward adoption of RS here is to maintain a fixed memory and
uniformly sample tasks from the task stream. Each task in the stream is assigned equal probability $\frac{n}{N}$ of being moved into the memory buffer, where $n$ is the memory capacity size and $N$ is the total number of tasks seen so far.  However, it is not suitable for the practical scenarios previously described, with two major shortcomings: (1) the task distribution in memory can be skewed when the input task stream is highly imbalanced in our setting. This leads to under-representation of the minority domains; (2) the importance of each task varies as some domains are more difficult to learn than others. This factor is also not taken into account with RS.

\begin{figure} 
     \centering
     \begin{subfigure}[b]{0.48\textwidth}
         \centering
         \includegraphics[width=\textwidth]{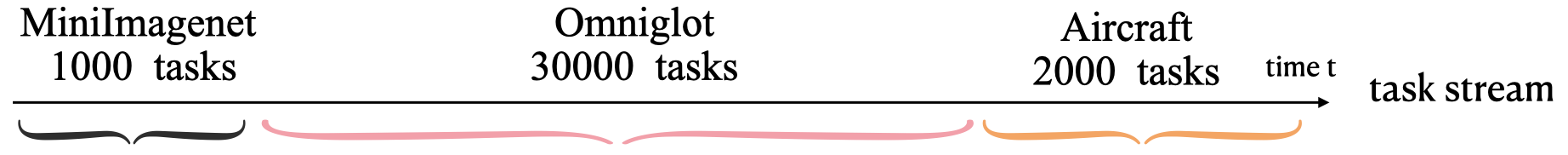}

     \end{subfigure}
     \hfill
     \begin{subfigure}[b]{0.21\textwidth}
         \centering
         \includegraphics[width=\textwidth]{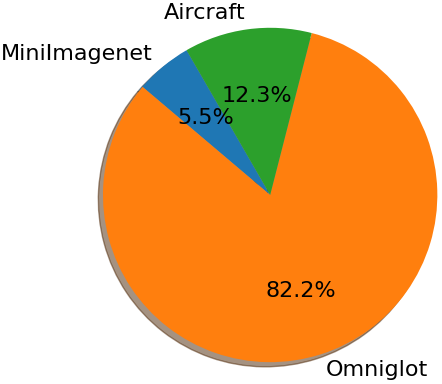}
         \caption{reservoir sampling}
     \end{subfigure}
          \begin{subfigure}[b]{0.26\textwidth}
         \centering
         \includegraphics[width=\textwidth]{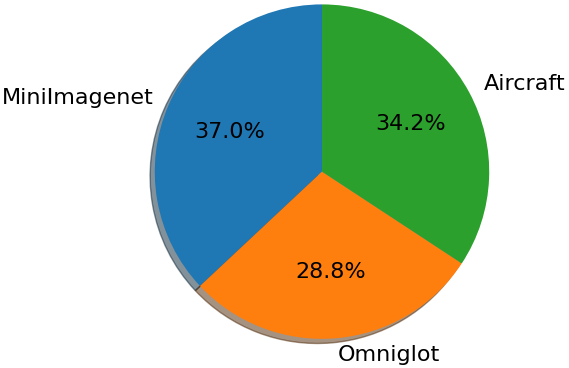}
         \caption{proposed memory management}
     \end{subfigure}
     \caption{An example of (a) reservoir sampling and (b) proposed memory management method jointly considering domain distribution and difficulty when meta learning on task stream from three latent domains.}
    \label{fig:memorymanage}
\end{figure}

To address the above issues, we propose to first detect domain change in the input task stream to associate each task with a latent domain label. We then present a new mechanism, called \textit{Memory Management with Domain Distribution and Difficulty Awareness} by utilizing the associated latent domain label with each task.  For simple illustration, we construct an imbalanced input task stream from Miniimagenet, Omniglot and Aircraft as shown in Figure \ref{fig:memorymanage}.
 Evidently, the resulting distribution of stored tasks with RS is highly imbalanced  and dramatically influenced 
by the input task stream distribution. In contrast, our memory management mechanism balances the three domain proportions by jointly considering domain distribution and difficulty. 

\textbf{Model Summary}: We first illustrate on our domain change detection component in Section \ref{sec:domaincluster}, which is used for (1) managing and balancing tasks in the memory buffer by incorporating the task difficulty (defined in Section \ref{sec:importancesampling}) to determine whether the new incoming task should be moved into the memory and which old task should be moved out of memory in Section \ref{sec:memorysampling}; (2) adaptive sampling tasks from memory buffer during meta training by dynamically adjusting the sampling probability of each task in the memory according to the task gradient for mitigating catastrophic forgetting in Section \ref{sec:importancesampling}.

\subsection{Online Domain Change Detection} \label{sec:domaincluster}

 Online domain change detection is a difficult problem due to: (1) few shot tasks are highly diverse within a single domain; (2) there are varying degree of variations at domain boundaries across the sequence. In our initial study,  we found that it is inadequate to set a threshold on the change of mini-batch task loss value for detecting domain change.  We thus construct a low dimensional projected space and perform online domain change detection on this space.

\paragraph{Projected space} Tasks $\mathcal{T}_t$  are mapped into a common space $\bm{o}_t = f_{\thetab_t}({\mathcal{T}_t^{train}})  = \frac{1}{K} \sum_{k=1}^{K} f_{\thetab_t}(\{\bm{x}^k \})$ where $K$ is the number of training data and $f_{\thetab_t}$ is the CNN embedding network. The task embedding could be further refined by incorporating the image labels, e.g., concatenating the word embedding of the image categories with image embedding. We leave this direction as interesting future work. To reduce the variance across different few shot tasks and capture the general domain information, we compute the exponential moving average of task embedding $ \bm{O}_{t}$ as $ \bm{O}_{t} = \alpha \bm{o}_t + (1-\alpha)\bm{O}_{t-1}$, where the constant $\alpha$ is the weighting multiplier which encodes the relative importance between current task embedding and past moving average. A sliding window stores the past $m$ ($m$ is a small number) steps moving average, $\bm{O}_{t-1}, \bm{O}_{t-2}, \cdots, \bm{O}_{t-m}$, which are used to form the low dimensional projection vector $\zb_t$, where the $i$-th dimensional element of $\zb_t$ is the distance between $\bm{o}_t$ and $\bm{O}_{t-i}$, $d(\bm{o}_t, \bm{O}_{t-i})$. The projected $m$ dimensional vector $\zb_t$ captures longer context similarity information spanning across multiple consecutive tasks. 

\textbf{Online domain change detection} At each time $t$, we utilize the above constructed projected space for online domain change detection. Assume we have two windows of projected embedding of previous tasks $\mathcal{U}^B = \{\zb_{t-2B}, \zb_{t-2B+1}, \cdots, \zb_{t-B-1} \}$ with distribution $Q$ and $\mathcal{V}^{B} = \{\zb_{t-B}, \zb_{t-B+1}, \cdots, \zb_t \}$ with distribution  $R$, where $B$ is the window size. In other words, $\mathcal{V}^{B}$ represents the most recent window of projection space (test window) and $\mathcal{U}^B$ represents the projection space of previous window (reference window). $\mathcal{U}^B$ and $\mathcal{V}^B$ are non-overlapping windows. For notation clarity and presentation convenience, we use another notation to denote the $\mathcal{U}^B = \{\ub_1, \ub_2, \cdots, \ub_B \}$ and $\mathcal{V}^B =  \{\vb_1, \vb_2, \cdots, \vb_B \}$, i.e., $\ub_i = \zb_{t-2B+i-1}$ and $\vb_i = \zb_{t-B+i-1}$. Our general framework is to first measure the distance between the two distributions $Q$ and $R$, $d(Q, R)$; then, by setting a threshold $b$, the domain change is detected when $d(Q, R)>b$. Here, we use  Maximum Mean Discrepancy (MMD) to measure the distribution distance. Following \cite{li2018scan}, the MMD distance between $Q$ and $R$ is defined as:

\begin{equation}
    \text{MMD}[\mathcal{F}, Q, R] := \sup\limits_{f\in \mathcal{F}} \{\mathbb{E}_{\ub \sim Q} [f(\ub)] - \mathbb{E}_{\vb \sim R} [f(\vb)]  \}
\end{equation}

U-statistics \cite{JMLRgretton12a} can be used for estimating ${\text{MMD}}^2$:

\begin{equation}
    W^B_t = \text{MMD}^2[\mathcal{U}^{B}, \mathcal{V}^{B}] = \frac{1}{B(B-1)} \sum_{i\neq j}^{B} h(\ub_i, \ub_j, \vb_i, \vb_j)
\end{equation}
and $h(\cdot)$ is defined as:

\begin{equation}
    h(\ub_i, \ub_j, \vb_i, \vb_j) = k(\ub_i, \ub_j) + k(\vb_i, \vb_j) - k(\ub_i,  \vb_j) - k(\ub_j,  \vb_i)
\end{equation}

where $k(\cdot, \cdot)$ is RKHS kernel.  In this paper,  we assume
RBF kernel $k(x, x^{\prime}) = exp(-||x- x^{\prime}||^2/2 \sigma^2) $ is used. 

The detection statistics at time $t$ is $W^B_t$.
If $Q$ and $R$ are close, $W^B_t$ is expected to be small, implying small probability of existence of domain change. If $Q$ and $R$ are significantly different distributions, $W^B_t$ is expected to be large, implying higher chance of domain shift. Thus, $W^B_t$ characterizes the chance of domain shift at time $t$. 
 We then test on the condition of $W^B_t>b$ to determine whether domain change occurs, where $b$ is a threshold. Each task $\mathcal{T}_t$ is associated with a latent domain label $L_t$, $L_0=0$. If $W^B_t>b$, $L_t = L_{t-1}+1$, i.e., a new domain arrives (Note that the actual domain changes could happen a few steps ago, but for simplicity, we could assume domain changes occur at time $t$); otherwise, $L_t = L_{t-1}$, i.e., the current domain continues. We leave the more general case with domain revisiting as future work. How to set the threshold is a  non-trivial task and is described in the following.

\textbf{Setting the threshold}   Clearly, setting the threshold $b$ involves a trade-off between two aspects: (1) the probability of $W^B_t>b$
 when there is no domain change; (2) the probability of
$W^B_t>b$ when there is domain change. As a result, if the domain similarity and difficulty vary significantly, simply setting a fixed threshold across the entire training process is highly insufficient. In other words, adaptive threshold of $b$ is necessary.  Before we present the adaptive threshold method, we first show the theorem which characterizes the property of detection statistics $W^B_t$ in the following.

 \begin{algorithm}[H]
  \small
	\caption{Online Domain Change Detection (ODCD).}
	\label{alg:adaptthresh}
	\begin{algorithmic}[1]
		\REQUIRE stream of detection statistics $W^B_t$; constant $\rho$; desired quantile  (significance level) $\delta$; Initialize $\mu_0 = 0$ and $\mu_0^{(2)} = 0$
		\STATE \textbf{Function}  ODCD ($W^B_t$, $\rho$, $\delta$)
				\STATE $\db = False$; // indicator of domain shift
		    \STATE $\mu_t = (1 - \rho)\mu_{t-1} + \rho (W^B_t)^2$
		    \STATE $\mu_t^{(2)} = (1 - \rho)\mu_{t-1}^{(2)} + \rho (W^B_t)^4$
		    \STATE $\sigma_t = \sqrt{\mu_t^{(2)} - \mu_t^2}$
            \IF{ $W^B_t> \mu_t + \delta\sigma_t$}
            \STATE $\db = True$; //there is domain shift at time $t$ 
            \ENDIF
            \RETURN $\db$
       \STATE \textbf{EndFunction}
       
	\end{algorithmic}
\end{algorithm}

\begin{theorem}
Assume $\zb_i$ are drawn i.i.d. from $Q$. Suppose that $\mathbb{E}_{Q}||k(\zb, \cdot) ||^4 < \infty$. Set $\mu \overset{def}{=} \mathbb{E}_{Q} k(\zb, \cdot)$ and $K(\zb, \zb^{\prime}) \overset{def}{=} \langle k(\zb, \cdot) -\mu, k(\zb^{\prime}, \cdot) -\mu \rangle$. Suppose the eigenvalue  $\xi_l$ and eigenvectors $\phi_l^2$ of $K$ satisfy $\xi_l \geq 0$ and $\mathbb{E}_Q \phi_l^2  < \infty$  such that $K(\zb, \zb^{\prime}) = \sum_{l\geq 1} \xi_l \phi_l(\zb) \phi_l(\zb^{\prime})$ and $\langle \phi_l, \phi_{l^{\prime}} \rangle = \textbf{1}_{l=l^{\prime}}$. Then,

\begin{equation}
      W^B_t \overset{d}{\to}  \beta \sum_{l\geq 1} \xi_l Z_l^2 
\end{equation}

\end{theorem}

Where $\overset{d}{\to}$ means converge in distribution and $(Z_l)_{l\geq 1}$ is a collection of  infinite independent standard normal random variables and $\beta$ is a constant. The theorem and proof follow from \cite{theoremkernel, newCP}. We can observe that $W^B_t$ asymptotically follows a distribution formed by a weighted linear combination of independent normal distribution.  By Lindeberg’s central limit theorem \cite{vaart_1998}, it is reasonable to assume $W^B_t$ is  approximately Gaussian distribution. The problem is thus reduced to estimate its mean $\mu_t$ and $\sigma_t$. The adaptive threshold $b$, following from \cite{newCP}, can be estimated by  online approximation, $b = \mu_t + \delta\sigma_t$, where $\delta$ is a constant and set to be the  desired quantile of the normal distribution. This adaptive method for online domain change detection is shown in Algorithm \ref{alg:adaptthresh}.

\subsection{Memory Management with Domain Distribution and Difficulty Awareness}
\label{sec:memorysampling}

\begin{figure}[h]
\centering
\includegraphics[width=7.6cm]{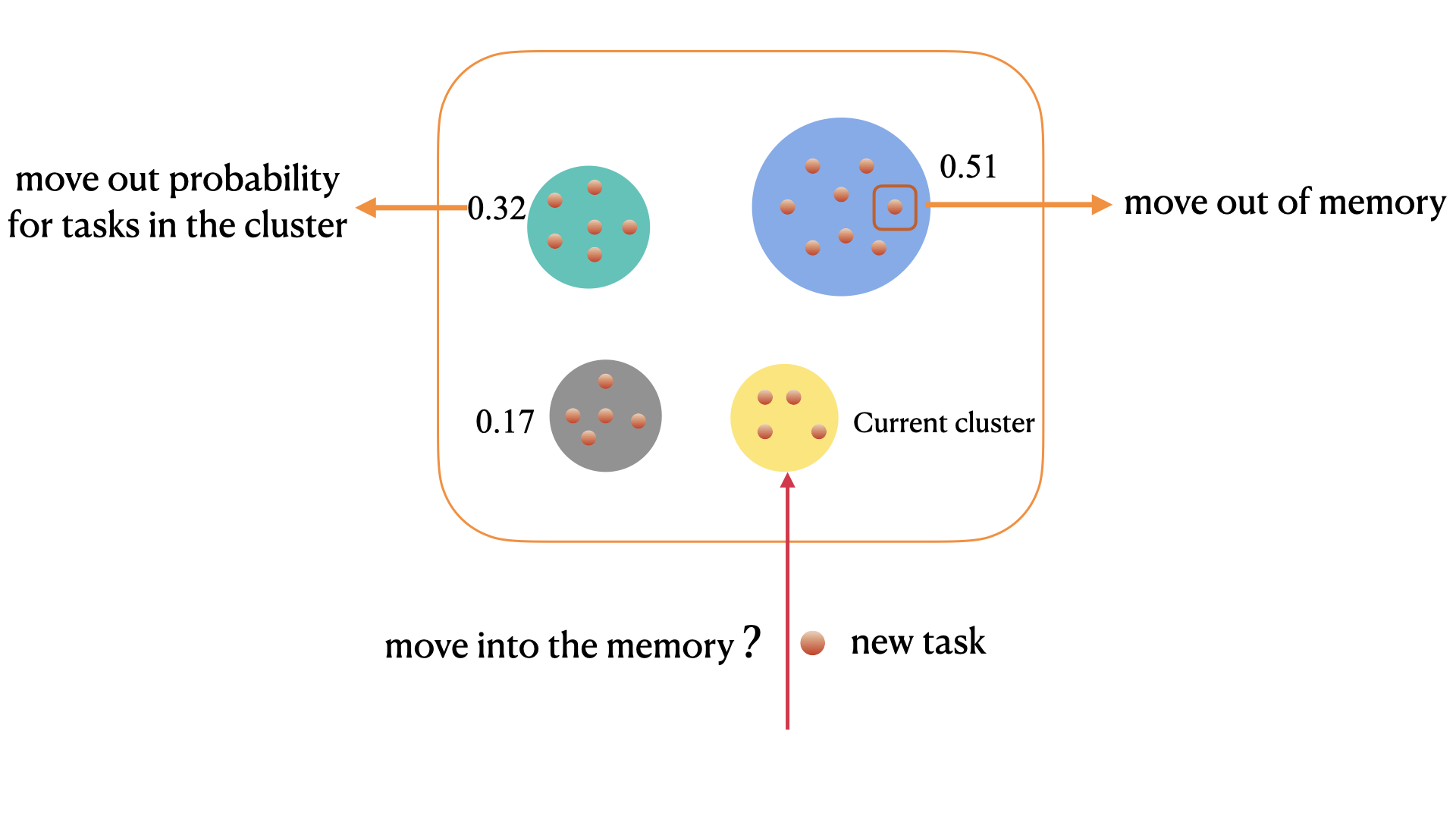}
\caption{Illustration on the memory management process. Each colored circle represents one cluster in the buffer and each dot denotes one task.}
 \label{fig:memory}
\end{figure}

In this section, we design the memory management mechanism for determining which task to be stored in the memory and which task to be moved out. The mechanism, named \textit{Memory Management with Domain Distribution and Difficulty Awareness} (M2D3), jointly considers the difficulty and distribution of few shot tasks in our setting. 
M2D3 first estimates the probability of the current task $\mathcal{T}_t$ to be moved into the memory.
The model will then determine the task to be moved out in the event that a new task move-in happens. 
To improve efficiency, we utilize the obtained latent domain information associated with each task (as described in previous section) to first estimate this move-out probability at cluster-level before sampling single task, as in Figure \ref{fig:memory}.

Here we define the notations involved in the following method description. Each task $\mathcal{T}_t$ in the memory is associated with a latent domain label $L_t$  and all the tasks with the same latent domain label form one cluster.  $\mathcal{M}_{i}$ denotes the cluster formed by all the tasks with latent domain label $i$ in memory $\mathcal{M}$, $n_{i} = |\mathcal{M}_{i}|$ denotes the number of tasks in $\mathcal{M}_{i}$ and $n = |\mathcal{M}|$ denotes the total number of tasks in memory, and $\mathcal{I}_i$ denotes the  importance score of  cluster $\mathcal{M}_i$. 

\textbf{Probability of new task moving into memory} 
When the new task $\mathcal{T}_t$ arrives, the chance of $\mathcal{T}_t$ being stored in memory is estimated, with the basic principle being the more incremental knowledge is brought by $\mathcal{T}_t$, the higher the probability of $\mathcal{T}_t$  being stored. This depends on the difficulty and prevalence of current latent domain. We propose an approach on top of this principle to estimate this probability. The score function of $\mathcal{T}_t$ is defined as: 

\begin{equation}
    S_{new} = (1-\frac{n_{L_{t}}}{n})\mathcal{I}_{t}^T
\end{equation}

Where $\mathcal{I}_{t}^{T}$ represents the importance for $\mathcal{T}_t$ , which is defined as the task-specific gradient norm in Section \ref{sec:importancesampling}. $n_{L_{t}}$ denotes the number of tasks of current latent domain cluster in memory buffer. $\frac{n_{L_{t}}}{n}$ denotes the prevalence of current latent domain in memory.  $\mathcal{I}_{i}$ represents the importance for cluster $\mathcal{M}_i$, which is defined as the cluster-specific gradient norm $G_i$ in Section \ref{sec:importancesampling} (The computation is shared and corresponding terms are computed only once.). The importance of in-memory tasks is defined as $ M_{s} = \frac{1}{L_t - 1} \sum_{i=1}^{L_t - 1} \frac{n_{i}}{n}\mathcal{I}_{i}$. The score function of in-memory tasks is defined as:

\begin{equation}
    S_{mem} =\frac{n_{L_{t}}}{n} M_{s}
\end{equation}

The probability of moving $\mathcal{T}_t$ into the memory is:

\begin{equation}\label{eq:in}
    P_{in} = \frac{e^{S_{new}}}{e^{S_{new}} + e^{S_{mem}}}
\end{equation}

This task selection mechanism maximizes the incremental knowledge of each task added into memory.

\begin{algorithm}[H]
  \small
	\caption{Memory Management with Domain Distribution and Difficulty Awareness (M2D3).}
	\label{alg:memory-manage}
	\begin{algorithmic}[1]
		\REQUIRE  mini-batch training tasks ${\mathcal{T}}_{t}$;  memory tasks $\mathcal{M}$; domain label $L_{t-1}$
    \STATE \textbf{Function} M2D3 $(\mathcal{M}, \mathcal{T}_t)$ 
    \STATE  calculate the probability $\mathcal{P}_{in}$ to move $\mathcal{T}_t$ into memory as Eq. \ref{eq:in}.
 calculate detection statistics of $W^B_t$
    \STATE $\db = $ ODCD($W^B_t$, $\rho$, $\delta$); detect domain change by Alg. \ref{alg:adaptthresh}.
    \IF {$\db$}
    \STATE $L_{t} = L_{t-1} +1$
    \STATE $\mathcal{M}_{L_{t}} = \{\}$
    \ENDIF
    \IF{memory $\mathcal{M}$ is not full}
        \STATE    $\mathcal{M}_{L_{t}} \leftarrow \mathcal{M}_{L_{t}} \cup \mathcal{T}_t$
    \ELSE
          \IF {${\mathcal{T}}_{t}$ is moved into memory by Eq. \ref{eq:in}}
    \STATE calculate the move-out probability for each cluster $\mathcal{P}_t^i$ and sample cluster $j$ according to  Eq. \ref{eq:score} and \ref{eq:sampleout}. 
    \STATE sample task from $\mathcal{M}_j$ to move out of memory.
    \STATE move $\mathcal{T}_t$ into memory $\mathcal{M}_{L_{t}} \leftarrow \mathcal{M}_{L_{t}} \cup {\mathcal{T}}_{t}$
    \ENDIF
    \ENDIF
  
    \RETURN updated memory buffer $\mathcal{M}$ 
    \STATE \textbf{EndFunction}
	\end{algorithmic}
\end{algorithm}

\textbf{Probability of existing tasks moving out of memory} 
To improve the efficiency of removing the tasks currently in memory, we perform a hierarchical sampling approach. We perform sampling first at cluster level before focusing on individual tasks, as shown in Figure \ref{fig:memory}. 
The estimated probability is correlated with both the size of each cluster in memory and its importance. The factor for each cluster $\mathcal{M}_i$ is defined as:

\begin{equation}\label{eq:score}
    \mathcal{A}_i \propto -(1-\frac{n_i}{n})\mathcal{I}_i
\end{equation}

The moving out probability for each cluster $\mathcal{M}_i$ at time $t$ is then defined as
\begin{equation}\label{eq:sampleout}
    \mathcal{P}_t^i = \frac{e^{\mathcal{A}_i}}{\sum_{i=1}^{i=L_t - 1} e^{\mathcal{A}_i}}
\end{equation}

The complete mechanism is summarized in Algorithm \ref{alg:memory-manage}.

\subsection{Adaptive Memory Task Sampling for Training} \label{sec:importancesampling}
\begin{figure}[h]
\centering
\includegraphics[width=8.0cm]{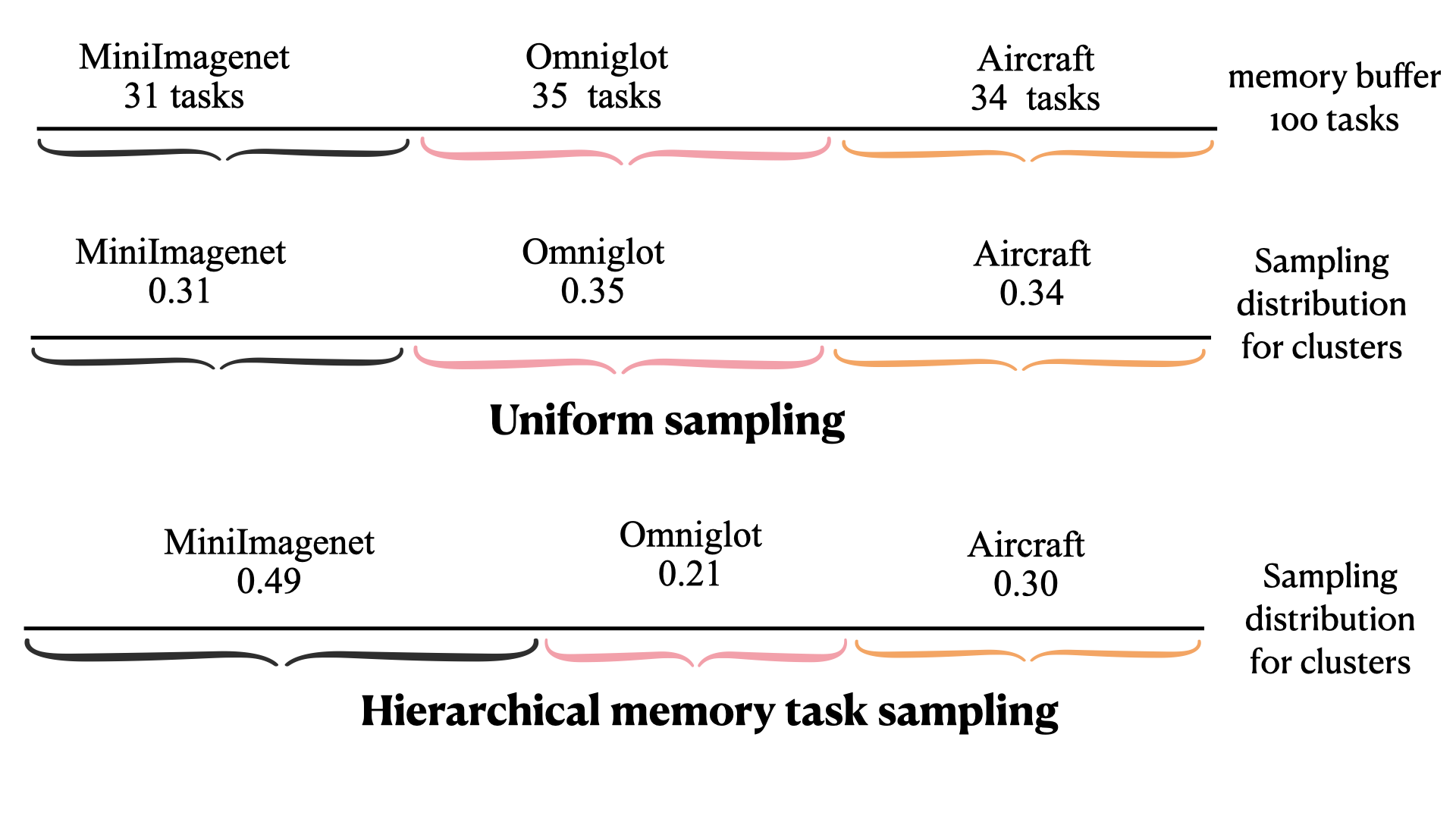}
\caption{A simple example of uniform task sampling and
our adaptive memory task sampling method for sampling tasks from memory buffer during meta training.}
 \label{fig:tasksampling}
\end{figure}

During meta training, a mini-batch of tasks are sampled from the memory and are jointly trained with current tasks to mitigate catastrophic forgetting. 
Direct uniform sampling tasks from memory incurs high variance, and results in unstable training \cite{khodak2019provable, NEURIPS2019_1dba5eed}. 
On the other hand, our intuition for non-uniform task sampling mechanism is that the tasks are not equally important for retaining the knowledge from previous domains. The tasks that carry more information are more beneficial for the model to remember previous domains, and should be sampled more frequently. To achieve this goal, we propose an efficient adaptive task sampling scheme in memory that accelerates training and reduces  gradient estimation variance. As shown in Figure \ref{fig:tasksampling}, the sampling probability of Miniimagenet and Aircraft are adjusted and increased based on the scheme suggesting the importance of these domains are higher than that of Omniglot for retaining knowledge.

 With the task specific loss function $\mathcal{L}_{\thetab}(\mathcal{T}_i) = P(\mathcal{T}^{test}_i|\thetab, \mathcal{T}^{train}_i)$. The optimization objective at time $t$ is defined as minimizing on the loss function of both the new tasks and memory tasks $\mathcal{H}(\thetab) = \mathcal{L}_{\thetab}(\mathcal{T}_t) + \sum \limits_{\mathcal{T}_i\in \mathcal{M}} \mathcal{L}_{\thetab}(\mathcal{T}_i)$. 

At time $t$, our proposed adaptive sampling mechanism assigns each task $\mathcal{T}_i \in \mathcal{M}$ a probability $q_i^t$ such that $\sum_{i=1} ^{i=n} q_i^t = 1$, we then sample $\mathcal{T}_{i_t}$ based on the distribution $\qb_t$ $= (q_1^t, q_2^t, \cdots, q_n^t)$. We temporally omit the subscript (superscript) $t$ for the following theorem for notation clarity.

\begin{theorem}\label{theorem:tasksample}
Let $\pb(\mathcal{T})$ be the distribution of the tasks in memory $\mathcal{M}$. Then,
\begin{equation}
    \mathbb{E}_{\pb(\mathcal{T})}  \nabla_{\thetab} \mathcal{L}_{\thetab}(\mathcal{T}) =  \mathbb{E}_{\qb(\mathcal{T})} [\frac{\pb(\mathcal{T})}{\qb(\mathcal{T})} \nabla_{\thetab} \mathcal{L}_{\thetab}(\mathcal{T}) ] = \Omega
\end{equation}

Let $\mathbb{V}_{\qb}[\Omega]$ denotes the covariance of the above estimator associated with $\qb$. Then, the trace of  $\mathbb{V}_{\qb}[\Omega]$ is minimized by the following optimal $\qb^{*}$

\begin{equation}
    \qb^{*}(\mathcal{T}) = \frac{\pb (\mathcal{T}) ||\nabla_{\thetab} \mathcal{L}_{\thetab}(\mathcal{T}) ||_2}{\int \pb (\mathcal{T}) ||\nabla_{\thetab} \mathcal{L}_{\thetab}(\mathcal{T}) ||_2}. 
\end{equation}

In particular, 
   if no prior information is available on task distribution, uniform sampling of tasks from memory is adopted and $\pb (\mathcal{T}) = \frac{1}{n}$,  $\qb^{*}(\mathcal{T}_i) = \frac{ ||\nabla_{\thetab} \mathcal{L}_{\thetab}(\mathcal{T}_i) ||_2}{ \sum_{j=1}^n ||\nabla_{\thetab} \mathcal{L}_{\thetab}(\mathcal{T}_j) ||_2}. $
     Thus, $w(\mathcal{T}_i) = \frac{\pb(\mathcal{T}_i)}{\qb(\mathcal{T}_i)}  =  \frac{1}{n \qb(\mathcal{T}_i)}$

\end{theorem}
Proof is provided in Appendix \ref{app:proof}. 
 The parameters are updated as:
\begin{equation}
    \thetab_{t+1} = \thetab_{t} - \eta w_i^{t} \nabla_{\thetab_{t}} \mathcal{L}_{\thetab}(\mathcal{T}_{i_t})
\end{equation}

Where $\eta$ is the learning rate, $w_i^t = \frac{1}{nq_i^t}$.
Similar to standard SGD analysis \cite{pmlr-v97-qian19b}, we define the convergence speed of meta training as the shrinkage of distance to optimal parameters $\thetab^{*}$ between two consecutive iterations $C = - \mathbb{E}_{\qb_t}  [||\thetab_{t+1} -\thetab^{*} ||_2^2 - ||\thetab_{t} -\thetab^{*} ||_2^2]$.
Following \cite{pmlr-v80-katharopoulos18a, alain2016variance}, it can be expressed as: 
\begin{equation}\label{eq:variance}
    C = 2\eta (\thetab_{t} -\thetab^{*}) \Omega - \eta^2 \Omega^{T}\Omega - \eta^2 \text{Tr}(\mathbb{V}_{\qb_t}[\Omega]  )
\end{equation}

 Theorem \ref{theorem:tasksample} illustrates the optimal task sampling distribution for reducing the gradient variance is proportional to the per-task gradient norm.  Minimizing the gradient variance (last term of RHS in Eq.\ref{eq:variance}) as in Theorem \ref{theorem:tasksample} also speeds up the convergence (maximize $C$) as a byproduct. However, it is computationally prohibitive to compute this distribution. We therefore propose efficient approximation to it.

By Section \ref{sec:domaincluster}, each memory task is associated with a latent cluster label. Utilizing this property, we can first sample $R$ (small) tasks from each cluster, then calculate the gradient norm for each cluster as $G_i$.

By doing so, the computational efficiency of the optimal task sampling distribution will be significantly improved. The sampling probability for each cluster is calculated as:

\begin{equation}\label{eq:outprob}
    \mathcal{Z}_t^i = \frac{n_i G_i}{\sum_{j=1}^{j=L_t}n_jG_j}
\end{equation}

The sampling scheme is to first sample cluster indexes from memory according to  Eq. \ref{eq:outprob}, then randomly sample tasks from the specified clusters.  We name this task sampling scheme as \textit{ada\textbf{P}tive m\textbf{E}mory \textbf{T}ask \textbf{S}ampling} (PETS).

Eq. \ref{eq:outprob} illustrates that the original sampling distribution of each cluster (measured by the frequency of each cluster in the memory buffer) is weighted by the corresponding importance of each cluster measured by the gradient norm $G_i$. In practice, the computational efficiency can be further improved by computing the sampling distribution every $s$ steps with the same distribution during each time interval. 
PETS is summarized in Algorithm \ref{alg:importance-sampling}.

 \begin{algorithm}[H]
  \small
	\caption{Adaptive Memory Task Sampling (PETS).}
	\label{alg:importance-sampling}
	\begin{algorithmic}[1]
		\REQUIRE A sequence of mini-batch training tasks ${\mathcal{T}}_{1}, {\mathcal{T}}_{2}, \dots, {\mathcal{T}}_{N}$;  memory buffer $\mathcal{M}$; model parameters $\thetab$;
		\FOR{$t = 1$ to $N$}
		   \FOR {each cluster $\mathcal{M}_j$ in $\mathcal{M}$}
		  \STATE sample mini-batch tasks from cluster $\mathcal{M}_j$ and calculate gradient norm $G_j$ for $\mathcal{M}_j$.
		 \ENDFOR
	\STATE calculate the task sampling distribution from each cluster $\mathcal{M}_j$ as in Eq. \ref{eq:outprob}.
    \STATE sample tasks $\mathcal{B}$ from  $\mathcal{M}$ according to the distribution $\mathcal{Z}_t$ as in Eq. \ref{eq:outprob}.
    \STATE update $\thetab$ by meta training on  $\mathcal{T}_t \cup \mathcal{B}$
        \STATE Memory tasks update
    $\mathcal{M} = \text{M2D3} (\mathcal{M}, \mathcal{T}_t)$
		\ENDFOR
	\end{algorithmic}
\end{algorithm}

\section{Related Work}

\paragraph{Meta Learning:}
Meta learning \cite{metanet} focuses on rapidly adapting to unseen tasks by learning on a large number of similar tasks. Representative works include \cite{matching16, protonet17, Edwards2017TowardsAN, finn17a, PMAML2018, pmlr-v48-santoro16, antoniou2018train, munkhdalai2017meta, learn2learn2016, SNAILICLR18, lee2019meta, Wang2020Bayesian, zhou2021metalearning, ravichandran2020fewshot, tokmakov2018learning, Zhang_2019_ICCV}, etc. All of these methods work on the simplified setting where task distributions are stationary during meta training. Completely different from these works, we focus on the more challenging setting where task distributions are non-stationary and imbalanced.

Online meta learning \cite{finn2019online}  stores all previous tasks in online setting to avoid forgetting with small number of tasks. \cite{jerfel2018reconciling} use Dirichlet process mixtures (DPM) to model the latent tasks structure and expand network. By contrast, ours focuses on mitigating catastrophic forgetting with single model when meta learning on imbalanced domain sequences with only limited access to previous domains.

Multi-domain meta learning \cite{triantafillou2019metadataset, tseng2020crossdomain, vuorio2019multimodal} assume tasks from all domains are available during meta training. We focus on the case that each domain in an imbalanced domain sequence sequentially arrives.

\paragraph{Continual Learning:}
Continual learning (CL) aims to maintain previous knowledge when learning on sequentially arriving data with distribution shift. Many works focus on mitigating catastrophic forgetting during the learning process. Representative works include \cite{lopezpaz2017gradient, AGEM19, riemer2018learning, yoon2018lifelong, EWC16, nguyen2017variational, ebrahimi2020adversarial,aljundi2018memory, IL2MCV, aljundi2019taskfree}, etc.  \textit{Continual few-shot learning} \cite{antoniou2020defining} (CFSL) focuses on remembering previously learned few-shot tasks in a single domain. 
 To our best knowledge, the replay-based approach to imbalanced streaming setting of continual learning has been only considered in \cite{aljundi2019gradient, imbalance2020, kim2020imbalanced}. Different from these works, which focus on learning on \textit{a small number of tasks} and aim to generalize to previous tasks, our work focuses on the setting where the model learns on \textit{a large number of tasks} with domain shift and imbalance, and aims to generalize to the \textit{unseen tasks} from previous domains without catastrophic forgetting instead of remembering on a specific task.

\paragraph{Incremental and Continual Few-shot Learning:}
Incremental few-shot learning \cite{gidaris2018dynamic, ren2018incremental, yoon2020xtarnet} aim to learn new categories while retaining knowledge on old categories within a single domain and assume access to the base categories is unlimited. This paper, by contrast,  requires good generalization to \textit{unseen} categories in previous domains and access to previous domains is limited. 

 Continual-MAML \cite{Caccia2020} aims for online fast adaptation to new tasks while accumulating knowledge on old tasks and assume previous tasks can be unlimited revisited.  MOCA \cite{harrison2020continuous} works in online learning and learns the experiences from previous data to improve sequential prediction. In contrast, ours focuses on generalizing to previous domain when learning on a large number of tasks with sequential domain shift and limited access to previous domains.

\section{Experiments}

 Our method is orthogonal to specific meta learning models and can be integrated into them seamlessly. 
For illustration, we evaluate our method on representative meta learning models including (1) gradient-based meta learning \textbf{ANIL} \cite{raghu2020rapid}, which is a simplified model of MAML \cite{finn17a}; (2) metric-based meta learning \textbf{Prototypical Network} (\textbf{PNet}) \cite{protonet17}. Extension to other meta learning models is straightforward.

\textbf{Baselines}: 
 (1) \textbf{sequential training}, which learns the latent domains sequentially without any external mechanism and demonstrates the model forgetting behavior; (2) \textbf{reservoir sampling (RS)} \cite{reservoir1985}; (3) \textbf{joint offline training}, which learns all the domains jointly in a multi-domain meta-learning setting; (4) \textbf{independent training}, which trains each domain independently. Among them, \textbf{joint offline training} and \textbf{independent training} serve as the performance upper bound.  
In addition,  since continual learning (CL) methods only 
apply to a small number of tasks, directly applying CL methods to our setting with large number of tasks (more than 40K) is infeasible. Instead, we combine several representative CL methods with meta learning base model.  We modify and adapt \textbf{GSS} \cite{aljundi2019gradient}, \textbf{MIR} \cite {aljundi2019online}, \textbf{AGEM} \cite{AGEM19} and \textbf{MER} \cite{riemer2018learning} to our setting and combine them with meta learning base models to serve as strong baselines. We denote these baselines as PNet-GSS, ANIL-GSS, etc.

\textbf{Proposed benchmark}
To simulate realistic imbalanced domain sequences, we construct a new benchmark and collect 6 domains with varying degree of similarity and difficulty, including \textbf{Quickdraw} \cite{quickdraw}, \textbf{AIRCRAFT} \cite{maji2013finegrained}, \textbf{CUB} \cite{WelinderEtal2010}, \textbf{Miniimagenet} \cite{matching16}, \textbf{Omniglot} \cite{Omniglot2011},  \textbf{Necessities} from Logo-2K+ \cite{wang2019logo2k}. We resize all images into the same size of $84\times84$. All the methods are compared for 5-way 1-shot and 5-way 5-shot learning. All the datasets are publicly available with more details provided in Appendix \ref{appendix:dataset}. We calculate the average accuracy on unseen testing tasks from all the domains for evaluation purpose.

\textbf{Implementation details} 
For ANIL-based \cite{raghu2020rapid} baselines, following \cite{antoniou2018train}, we use a four-layer CNN with 48 filters and one fully-connected layer as the meta learner. For PNet-based \cite{protonet17} baselines, we use a five-layer CNN with 64 filters of kernel size being 3 for meta learning. Following \cite{protonet17}, we do not use any fully connected layers for PNet-based models.  Similar architecture is commonly used in existing meta learning literature. We do not use any pre-trained network feature extractors which may contain prior knowledge on many pre-trained image classes, as this violates our problem setting that future domain knowledge is completely unknown. We perform experiments on different domain orderings, with the default ordering being Quickdraw, MiniImagenet, Omniglot, CUB, Aircraft and Necessities. To simulate imbalanced domains in streaming setting, each domain on this sequence is trained on 5000, 2000, 6000, 2000, 2000, 24000 steps respectively. In this setup, reservoir sampling will underrepresent
most domains. All experiments are averaged over three independent runs. More implementation details are given in Appendix \ref{app:detail}.

\subsection{Comparison to Baselines}
We compare our methods to the baselines. The memory maintains 300 batches (2) tasks. Results are shown in Table \ref{tab:protonetbaseline} and \ref{tab:ANILbaseline}. We can observe that our method significantly outperforms baselines by a large margin of  $5.21 \%$ for 5-shot learning and  $4.95\%$ for 1-shot learning with PNet-based model. For ANIL-based baselines, our method outperforms baselines by  $4.60 \%$ for 5-shot learning  and $2.19\%$ for 1-shot learning.  This shows the effectiveness of our method.

\subsection{Effect of Memory Capacity}
We explore the effect of memory capacity for the performance of  baselines and our method.
Table \ref{tab:ablation-memory1} and \ref{tab:ablation-memory2}  show the results with memory capacity (batches) of 200, 300 and 500 respectively. Our method significantly outperforms all the baselines in each capacity case.

\begin{table}
\centering
\caption{Comparisons with PNet-based baselines}
\begin{adjustbox}{scale=0.75,tabular= lccc,center}
\begin{tabular}{lrrrrrrr}
\toprule
 &&\multicolumn{2}{c}{5-Way 1-Shot}&  \multicolumn{2}{c}{5-Way 5-Shots}\\
Algorithm& & ACC & &ACC\\
\midrule  
PNet-Sequential && $31.82\pm 0.56$   && $48.21\pm 0.50$  \\
\midrule
PNet-RS && $34.68\pm 1.96$ && $53.69\pm 0.76$ \\
\midrule
PNet-GSS && $36.15\pm 1.59$ && $55.16\pm 0.72$\\
\midrule
PNet-AGEM && $34.07\pm 1.71$ && $52.61\pm 0.68$\\
\midrule
PNet-MIR &&$34.53\pm 1.45$  && $53.91\pm 0.56$\\
\midrule
PNet-MER &&$35.82\pm 1.69$ &&$54.28\pm 0.61$\\
\midrule
PNet-Ours && $\mathbf{41.10\pm 0.42}$ && $\mathbf{60.37\pm 0.32}$ \\
\midrule
Joint-training &&$52.96\pm 0.45$  && $68.56\pm 0.37$ \\
\midrule
Independent-training && $58.25\pm 0.36$  && $72.23\pm 0.29$ \\
\bottomrule
\end{tabular}
\label{tab:protonetbaseline}
\end{adjustbox}
\end{table}

\begin{table}
\centering
\caption{Comparisons with ANIL-based baselines}
\begin{adjustbox}{scale=0.75,tabular= lccc,center}
\begin{tabular}{lrrrrrrr}
\toprule
 &&\multicolumn{2}{c}{5-Way 1-Shot}&  \multicolumn{2}{c}{5-Way 5-Shots}\\
Algorithm& & ACC & &ACC\\
\midrule  
ANIL-Sequential &&  $30.68\pm 0.67$   && $41.39\pm 0.37$ \\
\midrule
ANIL-RS  && $32.11\pm 0.90$ && $48.72\pm 0.79$ \\
\midrule
ANIL-GSS && $31.78\pm 1.08$ && $48.93\pm 0.83$ \\
\midrule
ANIL-AGEM && $32.23\pm 1.21$ && $48.56\pm 0.91$  \\
\midrule
ANIL-MIR && $31.85\pm 0.97$ && $48.34\pm 0.72$  \\
\midrule
ANIL-MER &&$32.72\pm 1.06$ && $49.05\pm 0.96$ \\
\midrule
ANIL-Ours  && $\mathbf{34.91\pm 0.73}$ && $\mathbf{53.65\pm 0.56}$\\
\midrule
Joint-training &&  $52.37\pm 0.72$  && $66.21\pm 0.61$  \\
\midrule
Independent-training && $56.52\pm 0.57$  && $69.67\pm 0.53$  \\
\bottomrule
\end{tabular}
\label{tab:ANILbaseline}
\end{adjustbox}
\end{table}

\begin{table}
\centering
\caption{Effect of memory size for PNet-based baselines}
\begin{adjustbox}{scale=0.75,tabular= lccc,center}
\begin{tabular}{lrrrrrrr}
\toprule
 &&\multicolumn{2}{c}{5-Way 1-Shot}&  \multicolumn{2}{c}{5-Way 5-Shots}\\
Algorithm& & ACC & &ACC\\
\midrule
PNet-RS ($n= 200$) && $34.12\pm 1.12$  && $53.29\pm 0.42$ \\
PNet-Ours ($n= 200$) && $40.11\pm 0.73$  && $59.86\pm 0.27$ \\
\midrule
PNet-RS ($n= 300$) && $34.68\pm 1.96$ && $53.69\pm 0.76$ \\
PNet-Ours ($n= 300$) && $41.10\pm 0.42$ && $60.37\pm 0.32$ \\
\midrule
PNet-RS ($n= 500$) && $35.67\pm 0.82$  && $55.95\pm 0.79$\\
PNet-Ours ($n= 500$) && $41.82\pm 0.90$  && $61.05\pm 0.60$\\
\bottomrule
\end{tabular}
\label{tab:ablation-memory1} 
\end{adjustbox}
\end{table}

\begin{table}
\centering
\caption{Effect of memory size for ANIL-based baselines}
\begin{adjustbox}{scale=0.75,tabular= lccc,center}
\begin{tabular}{lrrrrrrr}
\toprule
 &&\multicolumn{2}{c}{5-Way 1-Shot}&  \multicolumn{2}{c}{5-Way 5-Shots}\\
Algorithm& & ACC & &ACC\\
\midrule
ANIL-RS ($n= 200$) && $31.03\pm 0.97$  &&  $45.96\pm 0.81$  \\
ANIL-Ours ($n= 200$) &&  $32.83\pm 0.71$ && $48.21\pm 0.61$ \\
\midrule
ANIL-RS ($n= 300$) && $32.11\pm 0.90$ && $48.72\pm 0.79$ \\
ANIL-Ours ($n= 300$) && $34.91\pm 0.73$ && $53.65\pm 0.56$\\
\midrule
ANIL-RS ($n= 500$) && $39.35\pm 0.76$ && $53.86\pm 0.68$ \\
ANIL-Ours ($n= 500$) && $42.79\pm 0.67$ && $59.23\pm 0.49$\\
\bottomrule
\end{tabular}
\label{tab:ablation-memory2} 
\end{adjustbox}
\end{table}

\subsection{Effect of Domain Ordering}

We also compare to other two orderings: \textbf{Necessities, CUB, Omniglot, Aircraft, MiniImagenet, Quickdraw}; and \textbf{Omniglot, Aircraft, Necessities, CUB, Quickdraw, MiniImagenet}. The results are shown in Appendix \ref{app:moreresults}. In all cases, our method substantially outperforms the baselines.

\subsection{Effect of Different Ratios of Domains}
To explore how the different domain ratios affect the model performance, we did another set of experiments with iterations of 4K, 4K, 3K, 4K, 4K, 22K steps on each domain respectively. The results are shown in Table \ref{tab:differentratio} in Appendix.

\subsection{Effect of Domain Revisiting}
To investigate the effect of domain revisiting on baselines and our method, we perform experiment on the domain sequence with domain revising of Quickdraw. The details and results are shown in Table \ref{tab:domainrevisit} in Appendix. We currently assume that there is no domain-revising, properly handling domain-revisiting is left as interesting future work.

\subsection{Ablation Study}

\textbf{Effect of memory management mechanism}
To verify the effectiveness of M2D3 proposed in section \ref{sec:memorysampling}, Table \ref{tab:gradientvariance} in Appendix shows the experiments with simple reservoir sampling without M2D3 (PNet-RS) and with M2D3 (PNet-Ours (without PETS)) respectively. Our method with M2D3 significantly outperforms baseline by $4.1\%$  and $4.2\%$ respectively. The memory proportion for each latent domain is shown in Figure \ref{fig:memorybalance} in Appendix. For RS baseline, the memory proportion for each domain is highly imbalanced. On the contrary, our memory management mechanism enables the memory proportion for each domain is relatively balanced, demonstrating the effectiveness of our method.

\textbf{Effect of PETS}
To verify the effectiveness of PETS proposed in section \ref{sec:importancesampling}, we compare the gradient variance with uniform sampling and our adaptive task sampling method, the gradient variance during training is shown in Figure \ref{fig:gradientvariance} in Appendix. We can see that our adaptive task sampling achieves much less gradient variance especially when training for longer iterations. Table \ref{tab:gradientvariance} in Appendix shows that with PETS, the performance is improved by more than $2.2\%$ and $2.4\%$ for 1-shot and 5-shot learning respectively.

\section{Conclusion}
This paper addresses the forgetting problem when meta learning on non-stationary and imbalanced task distributions. To address this problem, we propose a new memory management mechanism to balance the proportion of each domain in the memory buffer. Also, we introduce an efficient adaptive memory task sampling method to reduce the task gradient variance. Experiments demonstrate the effectiveness of our proposed methods.
For future work, it would be interesting to meta learn the proportional of each domain automatically. 

\textbf{Acknowledgements}
    This research was supported in part by NSF through grants IIS-1910492.

{\small
\bibliographystyle{ieee_fullname}
\bibliography{egbib}
}

\clearpage
\appendix

\textbf{\Large{Appendix}}

\section{Dataset Details} \label{appendix:dataset}

\paragraph{Miniimagenet \cite{matching16}} A subset dataset from ImageNet with 100 different classes, each class with 600 images.  The meta train/validation/test splits are  64/16/20 classes respectively, following the same splits of \cite{Ravi2017}. 

\paragraph{Omniglot \cite{Omniglot2011}} An image dataset handwritten characters from 50 different alphabets, with each class of 20 examples, following the same setup and data split in \cite{matching16}. 

\paragraph{CUB \cite{WelinderEtal2010}} A dataset consisting of 200 bird species. Following the same split of \cite{chen2019closerfewshot}, the meta train/validation/test splits are of 100/50/50 classes respectively. 

\paragraph{AIRCRAFT \cite{maji2013finegrained}} An image dataset for aircraft models consisting of 102 categories, with 100 images per class. Following the split in \cite{vuorio2019multimodal}, the dataset is split into 70/15/15 classes for meta- training/validation/test.

\paragraph{Quickdraw \cite{quickdraw}} An image dataset consisting of 50 million black-and-white drawings with 345 categories. Following \cite{Lee2020Learning}, the dataset is  split into 241/52/52 classes for
meta-training/validation/test.

\paragraph{Necessities} \textit{\textbf{Necessities}} Logo images from the large-scale publicly available dataset Logo-2K+ \cite{wang2019logo2k}. The dataset is randomly split into 100/41/41 classes for meta- training/validation/test.

\section{Implementation Detail}\label{app:detail}
We use 750 evaluation tasks from each domain for meta testing. $m = 5$ for constructing the projected space. $\delta = 1.64$ (corresponds to confidence level of $95 \%$) and window size $B=100$ for domain change detection. The meta batch size (number of training tasks at each iteration) is 2.  Our implementation is based on the Torchmeta library \cite{deleu2019torchmeta}.

We approximate each $||\nabla_{\thetab} \mathcal{L}_{\thetab}(\mathcal{T})||_2 \sim ||\nabla_{x_i} \mathcal{L}_{\thetab}(\mathcal{T})||_2 = G_i$; where $x_i$ is the pre-activation of last layer output of the network as in \cite{pmlr-v80-katharopoulos18a}.

\section{Theorem proof}\label{app:proof}

\begin{proof}

Let $\mu =  \mathbb{E}_{\pb(\mathcal{T})}  \nabla_{\thetab} \mathcal{L}_{\thetab}(\mathcal{T})$

\begin{multline}
      Tr(\mathbb{V}_{\qb}[\Omega]) =  \mathbb{E}_{\qb(\mathcal{T})} [ (\frac{\pb(\mathcal{T})}{\qb(\mathcal{T})} \nabla_{\thetab} \mathcal{L}_{\thetab}(\mathcal{T}) - \mu) \\
  (\frac{\pb(\mathcal{T})}{\qb(\mathcal{T})} \nabla_{\thetab} \mathcal{L}_{\thetab}(\mathcal{T}) - \mu)^T  ] = \mathbb{E}_{\qb(\mathcal{T})} [|| \frac{\pb(\mathcal{T})}{\qb(\mathcal{T})} \nabla_{\thetab} \mathcal{L}_{\thetab}(\mathcal{T})||_2^2 ] - ||\mu||_2^2
\end{multline}

 By Jensen's inequality:
  \begin{multline}
       \mathbb{E}_{\qb(\mathcal{T})} [|| \frac{\pb(\mathcal{T})}{\qb(\mathcal{T})} \nabla_{\thetab} \mathcal{L}_{\thetab}(\mathcal{T})||_2^2 ] \geq  \mathbb{E}_{\qb(\mathcal{T})} [|| \frac{\pb(\mathcal{T})}{\qb(\mathcal{T})} \nabla_{\thetab} \mathcal{L}_{\thetab}(\mathcal{T})||_2]^2 \\
       =  (\mathbb{E}_{\pb(\mathcal{T})}  [||\nabla_{\thetab} \mathcal{L}_{\thetab}(\mathcal{T})||_2])^2
  \end{multline}

The equality holds at $
    \qb^{*}(\mathcal{T}) = \frac{\pb (\mathcal{T}) ||\nabla_{\thetab} \mathcal{L}_{\thetab}(\mathcal{T}) ||_2}{\int \pb (\mathcal{T}) ||\nabla_{\thetab} \mathcal{L}_{\thetab}(\mathcal{T}) ||_2}. 
$ by plugging the above $\qb^{*}(\mathcal{T})$ into the covariance expression.
\end{proof}

\section{Additional Results}\label{app:moreresults}

\subsection{New ordering}

\textbf{Order 1: Omniglot, Aircraft, Necessities, CUB, Quickdraw, MiniImagenet }

To simulate imbalanced domains in streaming setting, each domain on this sequence is trained on 3000, 2000, 4000, 2000, 4000, 40000 steps respectively.

 \textbf{Order 2: Necessities, CUB, Omniglot, Aircraft, MiniImagenet, Quickdraw}

To simulate imbalanced domains in streaming setting, each domain on this sequence is trained on 6000, 2000, 6000, 3000, 3000, 24000 steps respectively.

\textbf{Order 1}. Table \ref{tab:memory1} shows the results. 
\textbf{Order 2}. Table \ref{tab:memory2} shows the results. 

\begin{table}
\centering
\caption{Effect of memory size  with order 1}
\begin{adjustbox}{scale=0.9,tabular= lccc,center}
\begin{tabular}{lrrrrrrr}
\toprule
 &&\multicolumn{2}{c}{5-Way 1-Shot}&  \multicolumn{2}{c}{5-Way 5-Shots}\\
Algorithm& & ACC & &ACC\\
\midrule  
PNet-RS ($n= 100$) && $36.82\pm 1.32$ && $54.83\pm 1.12$  \\
PNet-Ours ($n= 100$) && $42.35\pm 0.91$ && $59.85\pm 0.79$ \\
\midrule
PNet-RS ($n= 200$) && $37.30\pm 1.21$  && $55.23\pm 0.87$ \\
PNet-Ours ($n= 200$) && $43.61\pm 0.86$ && $61.32\pm 0.68$\\
\midrule
PNet-RS ($n= 300$) && $37.76\pm 1.19$  && $55.79\pm 0.91$ \\
PNet-Ours ($n= 300$) && $44.32\pm 0.83$ && $61.67\pm 0.57$\\
\midrule
PNet-RS ($n= 500$) && $38.82\pm 1.27$  && $55.95\pm 0.98$\\
PNet-Ours ($n= 500$) && $44.81\pm 0.63$ && $62.08\pm 0.61$\\
\bottomrule
\end{tabular}
\label{tab:memory1} 
\end{adjustbox}
\end{table}

\begin{table}
\centering
\caption{Effect of memory size  with order 2}
\begin{adjustbox}{scale=0.9,tabular= lccc,center}
\begin{tabular}{lrrrrrrr}
\toprule
 &&\multicolumn{2}{c}{5-Way 1-Shot}&  \multicolumn{2}{c}{5-Way 5-Shots}\\
Algorithm& & ACC & &ACC\\
\midrule  
PNet-RS ($n= 100$) && $43.08\pm 0.79$ &&  $57.97\pm 0.87$ \\
PNet-Ours ($n= 100$) && $46.67\pm 0.61$ && $62.14\pm 0.50$ \\
\midrule
PNet-RS ($n= 200$) && $43.36\pm 0.72$  && $58.23\pm 0.72$\\
PNet-Ours ($n= 200$) && $46.95\pm 0.52$ && $62.83\pm 0.58$\\
\midrule
PNet-RS ($n= 300$) && $44.16\pm 0.80$ && $58.65\pm 0.79$\\
PNet-Ours ($n= 300$) &&  $47.64\pm 0.45$ && $63.21\pm 0.46$\\
\midrule
PNet-RS ($n= 500$) && $45.29\pm 0.82$ && $59.36\pm 0.85$\\
PNet-Ours ($n= 500$) && $47.16\pm 0.49$  && $63.02\pm 0.46$ \\
\bottomrule
\end{tabular}
\label{tab:memory2} 
\end{adjustbox}
\end{table}

\begin{figure}
     \centering
     \begin{subfigure}[b]{0.5\textwidth}
         \centering
         \includegraphics[width=\textwidth]{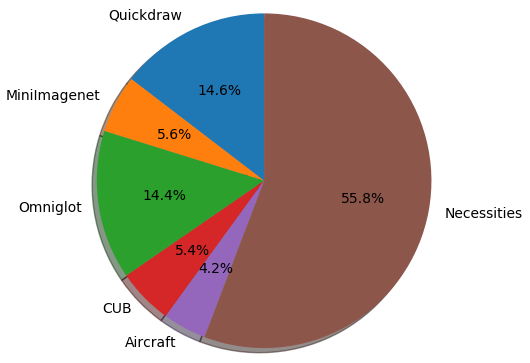}
         \caption{reservoir sampling}

     \end{subfigure}
     \hfill
     \begin{subfigure}[b]{0.5\textwidth}
         \centering
         \includegraphics[width=\textwidth]{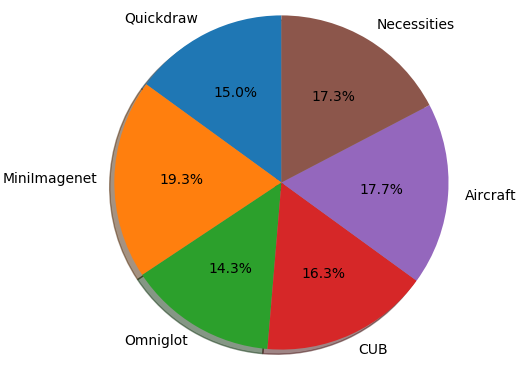}
         \caption{our memory management mechanism}
         \label{fig:balancedmemory}
     \end{subfigure}
     \caption{Results of different domain proportion in the memory of our memory management methods and reservoir sampling when meta learning on an imbalanced task stream from three latent domains.}
     \label{fig:memorybalance}
\end{figure}

\subsection{Effect of domain revisiting}
This section shows the results of effect of domain revisiting with domain ordering, \textbf{Quickdraw, MiniImagenet, Omniglot, CUB, Quickdraw, Aircraft, Necessities}. The domain Quickdraw is revisited. To simulate imbalanced domains in streaming setting, each domain on this sequence is trained on 5000, 2000, 6000, 2000, 3000, 2000, 24000 steps respectively.

\begin{table}
\centering
\caption{Comparisons with PNet-based baselines with domain revisiting}
\begin{adjustbox}{scale=0.75,tabular= lccc,center}
\begin{tabular}{lrrrrrrr}
\toprule
 &&\multicolumn{2}{c}{5-Way 1-Shot}&  \multicolumn{2}{c}{5-Way 5-Shots}\\
Algorithm& & ACC & &ACC\\
\midrule  
PNet-Sequential && $32.02\pm 0.50$ && $49.60\pm 0.45$   \\
\midrule
PNet-RS && $37.31\pm 1.56$ && $56.29\pm 1.35$ \\
\midrule
PNet-Ours &&$40.25\pm 0.98$  &&  $60.36\pm 0.83$\\
\midrule
Joint-training &&$52.96\pm 0.45$  && $68.56\pm 0.57$ \\
\midrule
Independent-training && $58.25\pm 0.36$  && $72.23\pm 0.29$ \\
\bottomrule
\end{tabular}
\label{tab:domainrevisit}
\end{adjustbox}
\end{table}

 \subsection{Effect of different ratios of domains}
 
 \begin{table}
\centering
\caption{Comparisons with PNet-based baselines with different imbalanced ratio of each domain}
\begin{adjustbox}{scale=0.75,tabular= lccc,center}
\begin{tabular}{lrrrrrrr}
\toprule
 &&\multicolumn{2}{c}{5-Way 1-Shot}&  \multicolumn{2}{c}{5-Way 5-Shots}\\
Algorithm& & ACC & &ACC\\
\midrule  
PNet-Sequential &&  $29.91\pm 0.71$  && $46.97\pm 0.65$ \\
\midrule
PNet-RS &&$34.97\pm 1.52$ && $54.79\pm 0.69$ \\
\midrule
PNet-GSS && $35.65\pm 1.28$ && $56.65\pm 0.81$\\
\midrule
PNet-AGEM &&$34.53\pm 1.36$ && $54.91\pm 0.73$ \\
\midrule
PNet-MIR && $35.09\pm 1.29$ && $54.56\pm 0.90$ \\
\midrule
PNet-MER && $35.16\pm 1.32$ && $55.71\pm 0.78$\\
\midrule
PNet-Ours && $40.57\pm 0.68$  && $61.53\pm 0.58$ \\
\midrule
Joint-training &&$52.96\pm 0.45$  && $68.56\pm 0.37$ \\
\midrule
Independent-training && $58.25\pm 0.36$  && $72.23\pm 0.29$ \\
\bottomrule
\end{tabular}
\label{tab:differentratio}
\end{adjustbox}
\end{table}

\subsection{Ablation Study}

\begin{figure}[h]
\centering
\includegraphics[width=9.0cm]{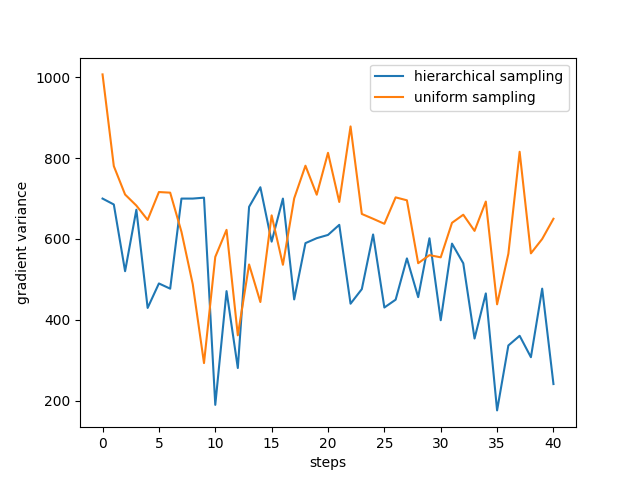}
\caption{Gradient variance comparison between uniform sampling  and PETS, each step (1000 iterations).}
 \label{fig:gradientvariance}
\end{figure}

\paragraph{Effect of PETS}
Figure \ref{fig:gradientvariance} shows the effect of sampling tasks with PETS.

\paragraph{Effect of memory management mechanism}
Table \ref{tab:gradientvariance} shows the effect of the proposed memory management mechanism. 

\begin{table}
\centering
\caption{Effect of Adaptive memory task sampling}
\begin{adjustbox}{scale=0.75,tabular= lccc,center}
\begin{tabular}{lrrrrrrr}
\toprule
 &&\multicolumn{2}{c}{5-Way 1-Shot}&  \multicolumn{2}{c}{5-Way 5-Shots}\\
Algorithm& & ACC & &ACC\\
\midrule  
PNet-RS && $34.68\pm 1.96$ && $53.69\pm 0.76$\\
\midrule  
PNet-Ours (without PETS) && $ 38.85\pm 0.79$   && $ 57.95\pm 0.67$  \\
\midrule
PNet-Ours (with PETS) && $41.10\pm 0.42$ && $60.37\pm 0.32$\\
\bottomrule
\end{tabular}
\label{tab:gradientvariance}
\end{adjustbox}
\end{table}

\end{document}